\documentclass[12pt]{article}

\usepackage[margin=1in]{geometry}
\usepackage{graphicx}
\usepackage{booktabs}
\usepackage{amsmath}
\usepackage{hyperref}
\usepackage[numbers]{natbib}
\usepackage{microtype}
\usepackage{setspace}
\usepackage{caption}
\usepackage{float}
\usepackage[T1]{fontenc}
\usepackage[utf8]{inputenc}
\usepackage{lmodern}
\usepackage{xcolor}

\hypersetup{
    colorlinks=true,
    linkcolor=blue,
    citecolor=blue,
    urlcolor=blue
}

\title{%
  \textbf{Forecasting Bacterial Antimicrobial Resistance Trends Using Machine Learning \\
  on WHO GLASS Surveillance Data: A Retrieval-Augmented Generation \\
  Approach for Policy Decision Support}
}

\author{
  Md Tanvir Hasan Turja \\
  Independent Researcher \\
  London, United Kingdom \\
  \url{https://github.com/TanvirTurja/amr-forecasting-rag}
}

\date{February 2026; v3 correction September 2026}

\begin{document}

\maketitle
\begin{center}
\fbox{\parbox{0.92\textwidth}{\small
\textbf{Correction (version 3, September 2026).} A subsequent audit of the benchmark
underlying this work identified four feature-construction defects: a lag feature equal to
the target itself on 8.7\% of test rows, specimen-crossed lags on 54\% of the remaining
rows, surveillance quality flags computed over the full 2020--2023 period, and full-panel
imputation. On the corrected panel the central performance claim of this version does not
hold: a one-year persistence forecast attains a test MAE of 5.79 percentage points,
significantly \emph{better} than XGBoost at 6.55 (paired country-cluster bootstrap,
$p = 0.001$), and the reported 85.3\% error reduction is an artifact of the defects.
Gradient boosting retains an advantage only on the 13\% of series lacking prior-year
observations. The audit procedure, corrected panel, and a persistence-first hybrid
selector are released at \url{https://github.com/TanvirTurja/amr-forecasting-rag}; a
complete reanalysis is under journal review. All remaining numbers and figures in this version are
reported as originally published and should be read subject to this correction.}}
\end{center}

\begin{abstract}
\textbf{Background:} Antimicrobial resistance (AMR) is a global health threat. While the WHO Global Antimicrobial Resistance and Use Surveillance System (GLASS) provides standardized data, population-level machine learning forecasting of resistance trends remains limited. Translating computational forecasts into policy also requires transparent interpretation mechanisms.

\textbf{Methods:} Surveillance data (2021--2023) comprising 5,909 observations across 44 countries and five WHO regions were processed. A rigorous temporal split prevented data leakage. Six models (Naive, Linear, Ridge, XGBoost, LightGBM, LSTM) were benchmarked to forecast one-year-ahead resistance rates using features including prior-year resistance and antibiotic consumption. Evaluation metrics (MAE, RMSE, sMAPE) were computed, with 95\% bootstrap confidence intervals generated for the MAE metric. A local Retrieval-Augmented Generation (RAG) system utilizing Gemma 4 was implemented to translate forecast findings into policy guidance grounded in retrieved WHO documents.

\textbf{Results:} XGBoost achieved the best predictive performance (test MAE = 6.13\% [95\% CI: 5.83--6.44]), representing a 85.3\% error reduction compared to the naive baseline (MAE = 41.79\%). SHAP (SHapley Additive exPlanations) analysis identified prior-year resistance as the dominant predictor (50.5\% gain), confirming strong autoregressive behavior. Regional forecast error tracked closely with surveillance coverage, ranging from 3.65\% in the European Region to 8.61\% in South-East Asia. The RAG pipeline generated accurate, source-attributed policy responses without fabricated citations.

\textbf{Conclusion:} Short-term AMR resistance rates exhibit strong temporal autocorrelation that can be accurately forecasted using gradient boosting. Coupling these forecasts with a hallucination-resistant RAG system provides a scalable, evidence-based decision-support framework for AMR governance.
\textbf{v3 correction:} a leakage audit of this benchmark found feature-construction
defects; on the corrected panel a persistence forecast (MAE 5.79) significantly
outperforms XGBoost (6.55), and the 85.3\% improvement claim is withdrawn (see
Correction).

\end{abstract}

\section{Introduction}

Antimicrobial resistance (AMR) is an escalating problem for global public health and economic stability. Evidence shows that the burden of drug-resistant bacterial infections falls disproportionately on low- and middle-income countries (LMICs), with Sub-Saharan Africa and South Asia experiencing the highest attributable mortality \cite{murray2022, dadgostar2019, tang2023}. Without coordinated interventions, the global health trajectory points toward 10 million annual AMR-associated deaths by 2050 \cite{oneill2016, dekraker2016}. To address this, the World Health Organization (WHO) established the Global Antimicrobial Resistance and Use Surveillance System (GLASS) to harmonise resistance data reporting internationally \cite{who2015gap, who2022glass, who2023manual}. While GLASS data has successfully quantified point-prevalence resistance and its correlation with antibiotic consumption \cite{ajulo2024, sugden2016}, its application for anticipatory, population-level forecasting remains constrained. This study focuses exclusively on bacterial AMR. Fungal resistance remains a long-neglected yet important component of the broader AMR framework, and incorporating fungal pathogens is a necessary direction for future modeling efforts \cite{fang2026}. For policymakers, assuming that next year's resistance will mirror last year's rate (a naive baseline) is insufficient. Because AMR rates can shift rapidly due to local stewardship interventions or consumption spikes, proactive governance requires complex models capable of capturing non-linear relationships between antibiotic usage and resistance.

Machine learning (ML) models can forecast epidemiological trends, yet existing AMR applications focus predominantly on classifying individual clinical isolates rather than predicting macro-level temporal trends \cite{sakagianni2023, kim2022}. Computational forecasts often suffer from the ``black box'' problem, lacking the interpretability required by public health stakeholders to formulate policy. Concurrently, while Large Language Models (LLMs) present a novel interface for querying complex datasets, their propensity to fabricate citations limits their utility in evidence-based medicine without constraints like Retrieval-Augmented Generation (RAG) \cite{lewis2020}.

The objective of this study was twofold: (1) to benchmark the forecasting accuracy of multiple ML architectures on WHO GLASS surveillance data to predict short-term national resistance rates, and (2) to design and evaluate a locally deployed RAG pipeline that grounds these computational forecasts in verifiable WHO policy documents.

\section{Methods}

\subsection{Dataset and Ethical Considerations}

Data spanning 2021 to 2023 were extracted from the WHO GLASS database and aggregated by country, WHO region, and year. As this study utilized publicly available, anonymized, and aggregated surveillance data, formal ethical approval and patient consent were not required. The consolidated dataset contained 5,909 unique observations (country, pathogen, antibiotic, year) across 44 nations. The target variable was the resistance percentage. This study is reported in accordance with the TRIPOD+AI (Transparent Reporting of a multivariable prediction model for Individual Prognosis Or Diagnosis + Artificial Intelligence) guidelines; a completed checklist is provided as supplementary material.

\subsection{Feature Engineering and Preprocessing}

Input features comprised year, pathogen identity, infection type, antibiotic name, antibiotic class, country or territory, WHO region, World Bank income group, total country observation count, high-quality surveillance flag, antibiotic consumption (defined daily doses per 1,000 inhabitants per day), prior-year antibiotic consumption (DID\_lag1), and resistance lag (the resistance percentage in the prior year for the identical pathogen-antibiotic-country cohort). Missing values in continuous features were imputed using the median rate for the specific pathogen-antibiotic pair. Median imputation was selected over mean imputation to prevent the estimates from being skewed by extreme resistance outliers (e.g., highly resistant localized outbreaks) within small sample cohorts, thereby preserving the central tendency without artificially inflating variance. High-cardinality categorical variables were encoded using target encoding, fitted exclusively on the training data.

A two-step temporal training strategy was enforced to prevent data leakage. In the first step, observations from 2021 served as the training set and 2022 data as the validation set, used solely to determine optimal early-stopping iterations for the gradient boosting models and the best training epoch for the LSTM. In the second step, final models were retrained on the combined 2021--2022 data using these fixed hyperparameters and evaluated on the held-out 2023 test set. Data prior to 2021 were deliberately excluded because the computation of the primary autoregressive feature (resistance lag-1) requires contiguous preceding-year data, which was inconsistently available or absent in the 2020 GLASS release due to pandemic-related reporting disruptions. Derived autoregressive features (e.g., rolling averages) were excluded to prevent mathematical target leakage. A rolling two-year resistance feature was present in the raw dataset, but it was found to be mathematically identical to the 1-year lag feature for the vast majority of observations due to data sparsity. To prevent multicollinearity, this rolling feature was entirely excluded from model training.

\subsection{Model Development and Benchmarking}

Six architectures were benchmarked: a Naive Baseline (predicting all test observations as a single constant, i.e., the final training-set resistance value, denoting $t$ as the last observed $t-1$), Linear Regression, Ridge Regression, XGBoost \cite{chen2016}, LightGBM \cite{ke2017}, and a Long Short-Term Memory (LSTM) neural network \cite{chimmula2020}. Gradient boosting frameworks have previously demonstrated strong predictive capacity for infectious disease time series \cite{alim2020, fang2022, ahn2023}. Hyperparameters for XGBoost and LightGBM were manually specified based on iterative validation-set performance, with early stopping to prevent overfitting. The LSTM architecture comprised two recurrent layers (hidden size = 64) with a dropout rate of 0.3, followed by a fully connected output layer. LSTM training utilized the Adam optimizer (learning rate = 0.001, weight decay = $1 \times 10^{-5}$) for up to 100 epochs with a batch size of 64, a ReduceLROnPlateau scheduler (patience = 5, factor = 0.5), and early stopping with a patience of 30 epochs based on validation MSE loss. All data processing and modelling were performed using Python 3.13 and PyTorch 2.6 on a single NVIDIA GeForce RTX 5060 Laptop GPU with CUDA acceleration. Traditional time-series models such as ARIMA were considered but ultimately excluded from the benchmark; partitioning national-level data by pathogen and antibiotic yields thousands of highly fragmented, short time-series (e.g., three temporal steps), rendering univariate autoregressive modeling computationally impractical and statistically unstable compared to cross-sectional gradient boosting.

\subsection{Evaluation and Interpretability}

Models were evaluated on the 2023 test set using Mean Absolute Error (MAE), Root Mean Squared Error (RMSE), symmetric Mean Absolute Percentage Error (sMAPE), and $R^{2}$. sMAPE was selected over the conventional MAPE metric to prevent division-by-zero errors when true resistance values are exactly 0\%, which occur commonly in the dataset. Statistical significance was established by generating 95\% bootstrap confidence intervals (2,000 iterations) for the MAE metric. To provide direction-aware interpretability, SHapley Additive exPlanations (SHAP) summary and dependence analyses were conducted on the best-performing model.

\subsection{Retrieval-Augmented Generation (RAG) Pipeline}

A RAG system was engineered to answer natural-language policy questions. A highly targeted knowledge base was constructed from six foundational WHO policy documents: (1) the GLASS Report 2021 \cite{who2022glass}, (2) the GLASS Report 2022 \cite{who2022glass}, (3) the WHO Global Action Plan on Antimicrobial Resistance \cite{who2015gap}, (4) the GLASS Manual for Antimicrobial Resistance Surveillance \cite{who2023manual}, (5) the WHO Bacterial Priority Pathogens List, 2024 \cite{who2024bppl}, and (6) the WHO Policy Guidance on Integrated Antimicrobial Stewardship Activities \cite{who2024stewardship}. This constrained document set was deliberately selected to minimize retrieval noise, prevent hallucination from tangential literature, and ensure the language model was grounded strictly in high-level, authoritative macro-policy. Documents were chunked and embedded using the all-MiniLM-L6-v2 sentence transformer and indexed in ChromaDB. For each query, the top five most relevant chunks were retrieved via hybrid BM25 keyword and dense vector search with Reciprocal Rank Fusion (RRF). These chunks, alongside the empirical XGBoost forecast metrics, were injected into a strict system prompt. The prompt explicitly prohibited fabricated citations, requiring the Gemma 4 (4B parameters) language model (served locally via Ollama) to synthesize answers grounded solely in the provided context.

\section{Results}

\subsection{Epidemiological Baselines and Data Complexity}

\begin{figure}[H]
\centering
\includegraphics[width=0.85\textwidth]{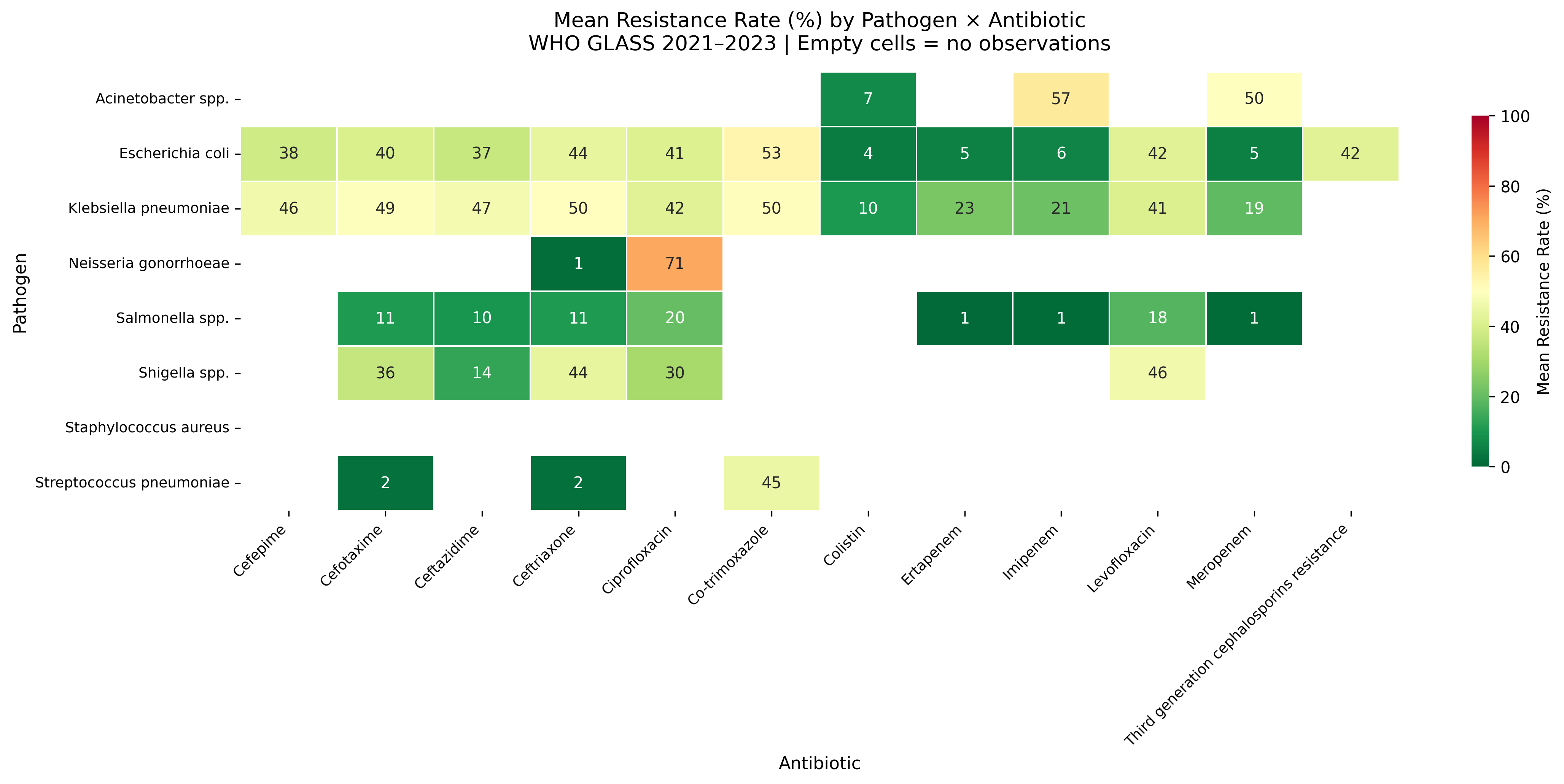}
\caption{Global resistance heatmap across priority pathogens. High variance shows the need for non-linear, pathogen-aware predictive models.}
\label{fig:heatmap}
\end{figure}

Prior to predictive modelling, an exploratory analysis of the WHO GLASS dataset revealed substantial heterogeneity in resistance rates across pathogens and regions. The resistance heatmap (Figure~\ref{fig:heatmap}) illustrates the baseline complexity of the forecasting task, with high-priority Gram-negative pathogens (e.g., \textit{Acinetobacter} spp., \textit{Klebsiella pneumoniae}) exhibiting severe resistance burdens particularly in under-resourced settings, necessitating non-linear modeling approaches capable of capturing pathogen-specific interaction effects.

\subsection{Forecasting Performance}

\begin{figure}[H]
\centering
\includegraphics[width=0.85\textwidth]{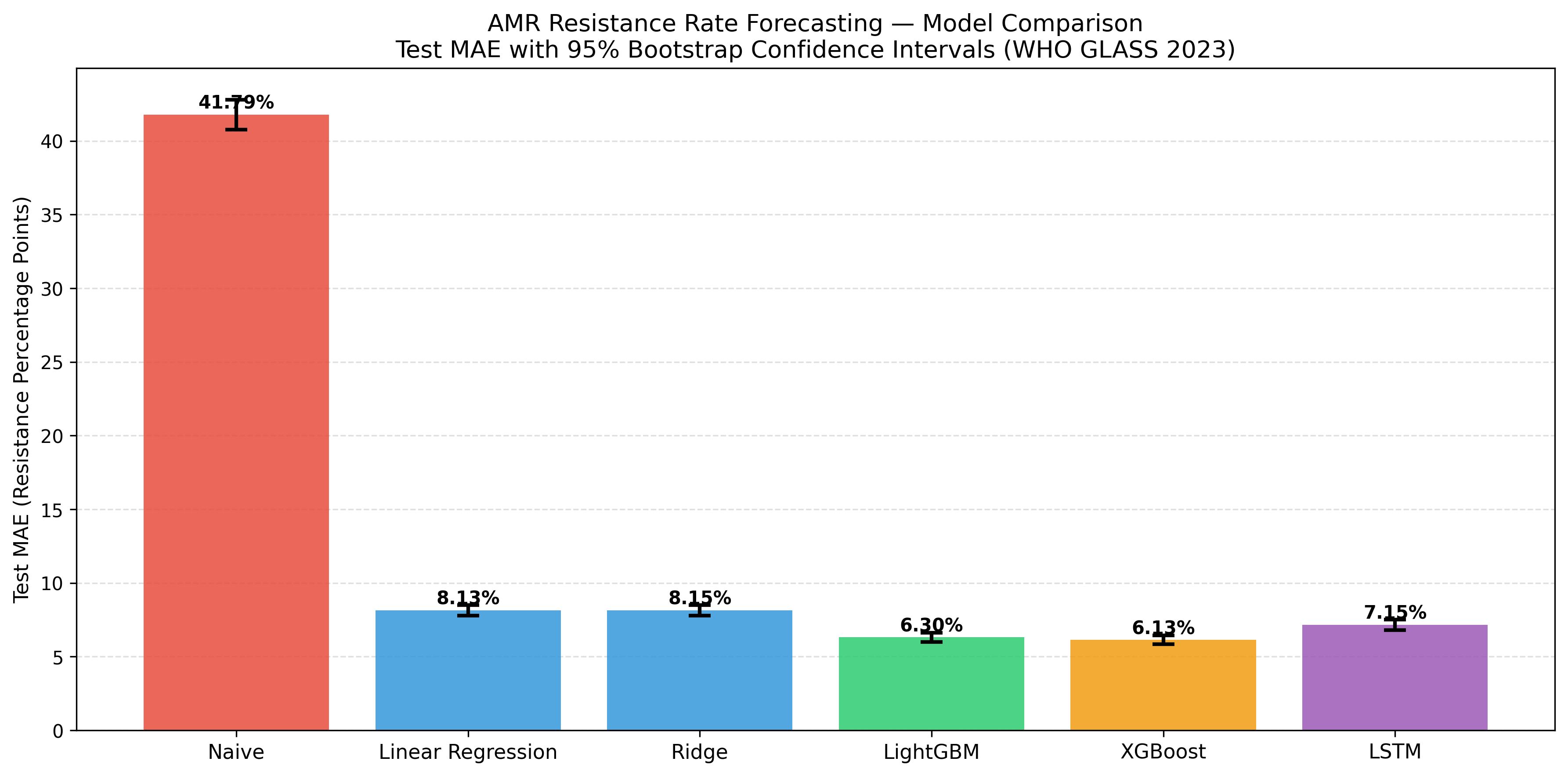}
\caption{Test MAE across models with 95\% bootstrap confidence intervals. Narrow overlap between XGBoost and LightGBM indicates statistical equivalence, while XGBoost significantly outperforms the LSTM given the limited temporal dataset.}
\label{fig:model_comparison}
\end{figure}

\begin{figure}[H]
\centering
\includegraphics[width=0.85\textwidth]{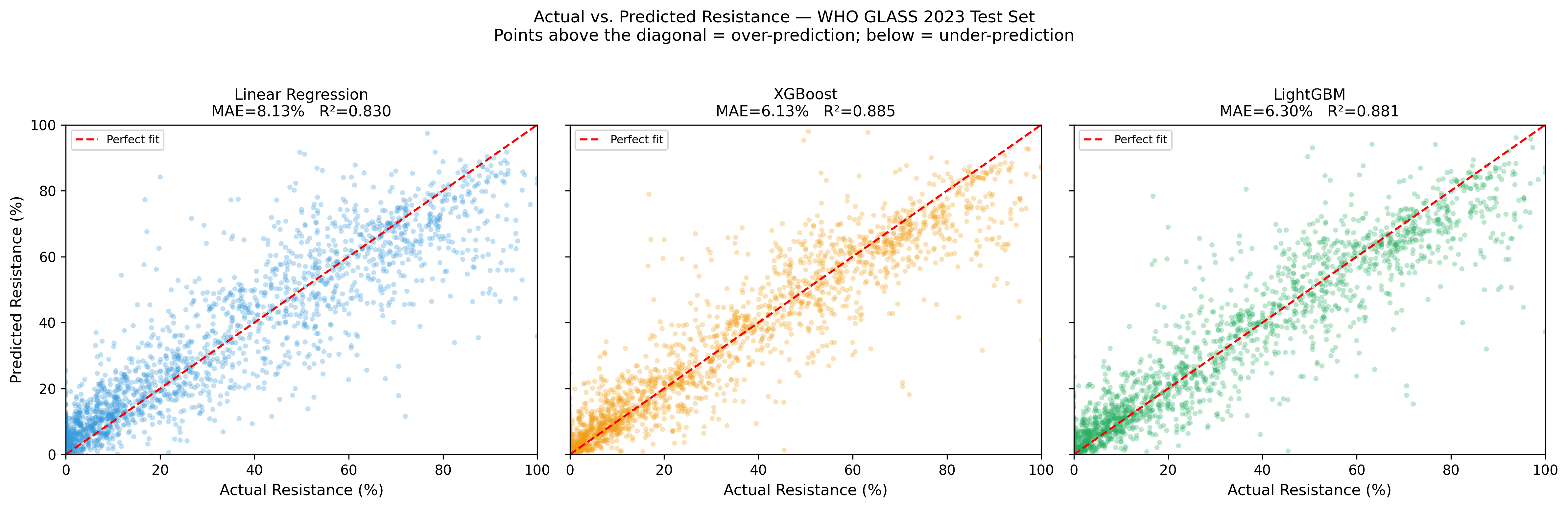}
\caption{Actual vs. predicted resistance percentages for three representative models (Linear Regression, XGBoost, LightGBM), demonstrating strong fit ($R^{2} \geq 0.83$) without systematic under-prediction of high-resistance events.}
\label{fig:actual_vs_predicted}
\end{figure}

XGBoost achieved the highest predictive accuracy on the held-out 2023 test set, recording an MAE of 6.13\% (95\% CI: 5.83--6.44) and an $R^{2}$ of 0.885. This represented a 85.3\% reduction in error compared to the Naive Baseline (MAE = 41.79\% [95\% CI: 40.77, 42.79]). Performance between the top three models was comparable; LightGBM recorded an MAE of 6.30\% (95\% CI: 6.00--6.63) and the LSTM recorded 7.15\% (95\% CI: 6.79--7.52). Linear and Ridge regression models yielded slightly higher errors at 8.13\% and 8.15\% MAE, respectively (Table~\ref{tab:results} and Figure~\ref{fig:model_comparison}). The parity between actual and predicted resistance percentages for three representative models confirms predictive accuracy across both low- and high-resistance cohorts (Figure~\ref{fig:actual_vs_predicted}). The bootstrap confidence intervals of XGBoost [5.83--6.44] and LightGBM [6.00--6.63] substantially overlap, indicating that their performance difference is not statistically significant. However, the XGBoost and LSTM confidence intervals [6.79--7.52] do not overlap, confirming a statistically significant advantage for the gradient boosting approach on this limited temporal dataset. With only three years of training data, the LSTM operates with a sequence length of one, which reduces it to a nonlinear regressor rather than a true temporal model. This architectural constraint means the LSTM cannot exploit long-range sequential dependencies, and its competitive performance likely stems from the expressivity of its fully connected output layers rather than recurrent dynamics. A fairer deep learning baseline under these conditions might be a standard multilayer perceptron (MLP); however, the LSTM was retained to establish a benchmark for future extensions where longer GLASS time series become available.

\begin{table}[H]
\centering
\caption{Model performance on the 2023 held-out test set. $^{a}$Predicting all test observations as a single constant (the final training-set resistance value).}
\label{tab:results}
\resizebox{\textwidth}{!}{%
\begin{tabular}{lccccc}
\toprule
\textbf{Model} & \textbf{Test MAE (\%)} & \textbf{95\% CI} & \textbf{Test sMAPE (\%)} & \textbf{Test RMSE (\%)} & \textbf{Test $R^{2}$} \\
\midrule
Naive Baseline$^{a}$   & 41.79          & [40.77, 42.79]   & 103.23 & 48.15 & ---   \\
Linear Regression      & 8.13           & [7.78, 8.49]     & 61.67  & 11.67 & 0.830 \\
Ridge Regression       & 8.15           & [7.79, 8.50]     & 61.64  & 11.67 & 0.830 \\
\textbf{XGBoost}       & \textbf{6.13}  & \textbf{[5.83, 6.44]} & \textbf{49.72} & 9.57 & \textbf{0.885} \\
LightGBM               & 6.30           & [6.00, 6.63]     & 50.55  & 9.76 & 0.881 \\
LSTM                   & 7.15           & [6.79, 7.52]     & 54.05  & 10.84 & 0.853 \\
\bottomrule
\end{tabular}%
}
\end{table}

\subsection{Feature Importance and Interpretability}

\begin{figure}[H]
\centering
\includegraphics[width=0.85\textwidth]{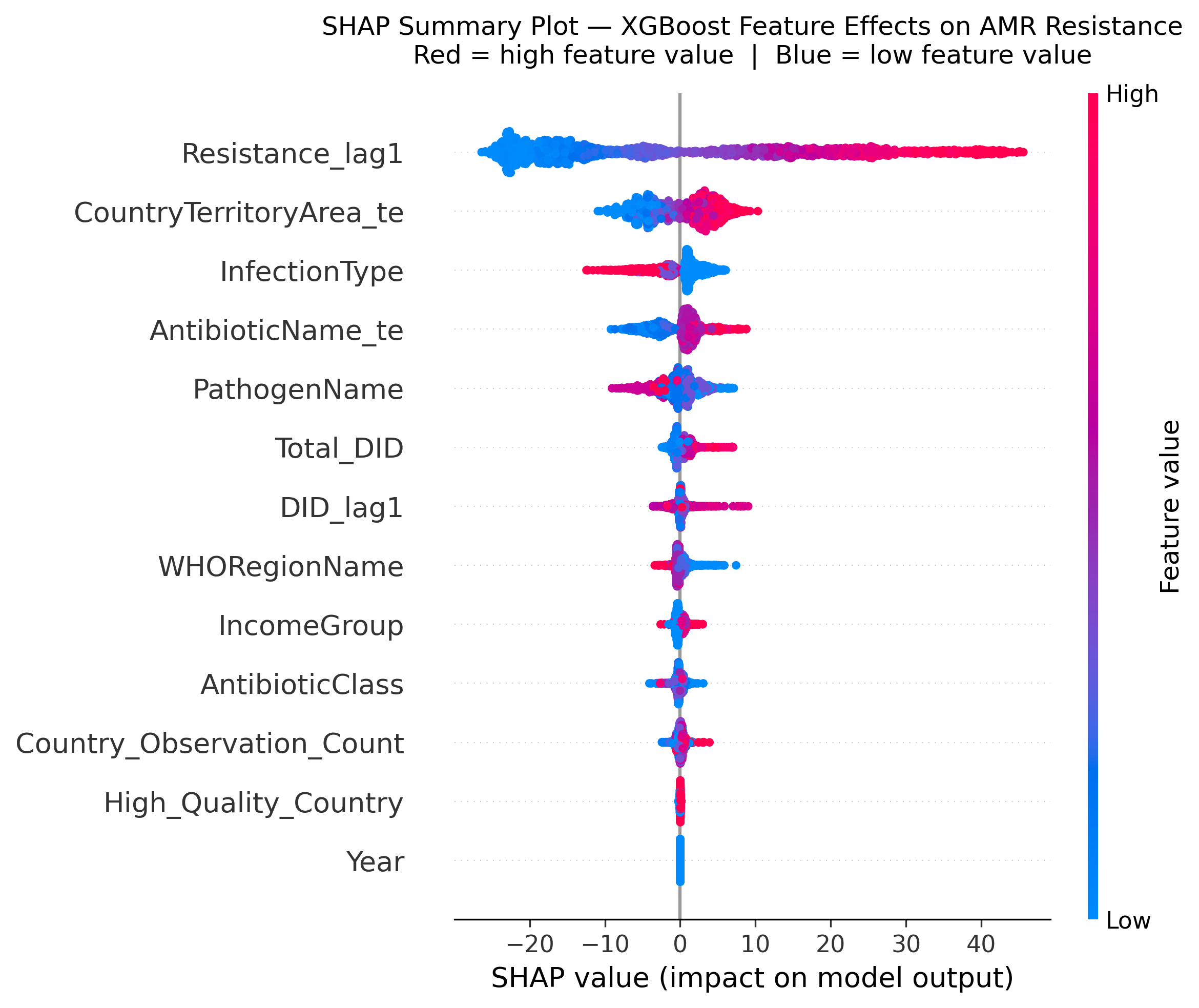}
\caption{SHAP summary plot for the XGBoost model. Features are ranked by overall impact on model output, providing direction-aware global interpretability.}
\label{fig:shap_summary}
\end{figure}

The prior-year resistance rate (lag-1) was the dominant predictor of future resistance. SHAP analysis of the XGBoost model confirmed that higher lag-1 values monotonically increased the predicted resistance output (Figure~\ref{fig:shap_summary}). Gain-based importance metrics corroborated this, with Resistance\_lag1 accounting for 50.5\% of the predictive weight, followed by Country/Territory (9.0\%), Antibiotic Name (6.7\%), Pathogen Name (6.0\%), Infection Type (5.3\%), and Income Group (5.1\%). Antibiotic consumption (Total\_DID) ranked ninth at 2.7\%.

\begin{table}[H]
\centering
\caption{XGBoost feature importances (gain-based, top 6 and rank 9).}
\label{tab:importance}
\begin{tabular}{clc}
\toprule
\textbf{Rank} & \textbf{Feature} & \textbf{Importance (\%)} \\
\midrule
1 & Resistance\_lag1 (prior year resistance) & 50.5 \\
2 & CountryTerritoryArea                    & 9.0  \\
3 & AntibioticName                          & 6.7  \\
4 & PathogenName                            & 6.0  \\
5 & Infection Type                          & 5.3  \\
6 & Income Group                            & 5.1  \\
9 & Antibiotic consumption (Total\_DID)     & 2.7  \\
\bottomrule
\end{tabular}
\end{table}

\begin{figure}[H]
\centering
\includegraphics[width=0.85\textwidth]{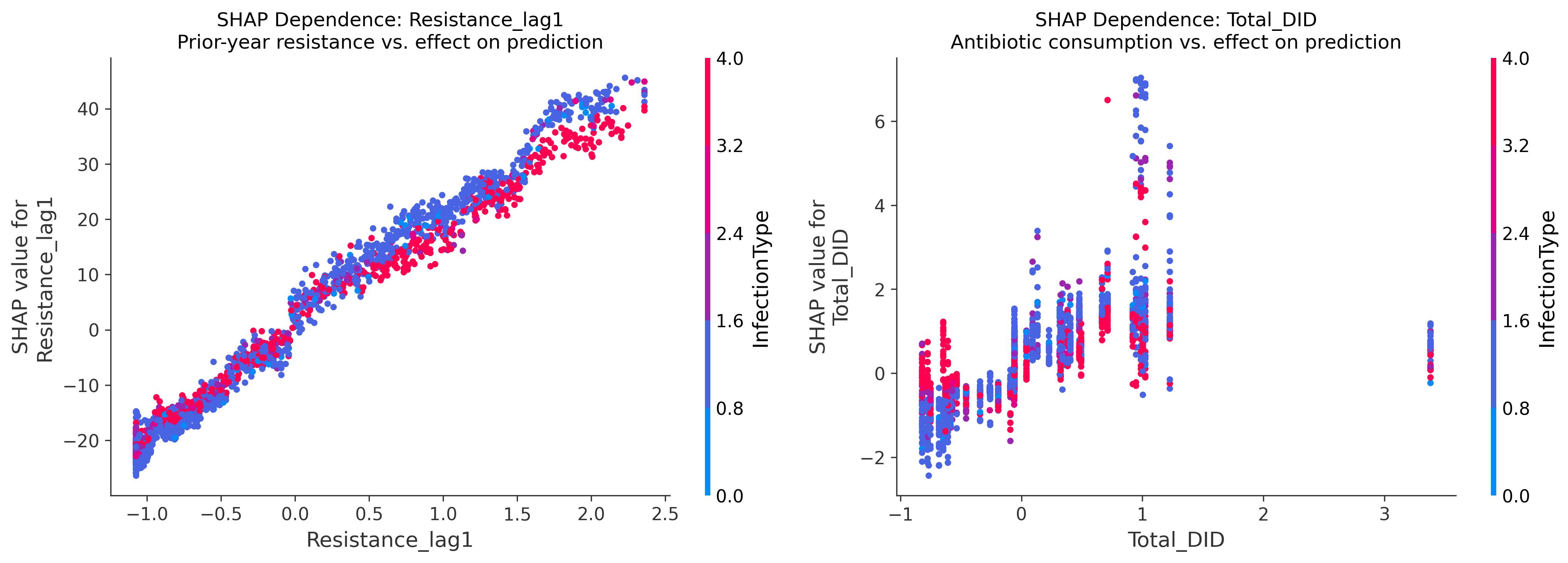}
\caption{SHAP dependence plots showing the non-linear impact of resistance lag and antibiotic consumption on model predictions.}
\label{fig:shap_dependence}
\end{figure}

The SHAP dependence plots further unravel these non-linear effects (Figure~\ref{fig:shap_dependence}). The model isolated the epidemiological impact of antibiotic consumption (defined daily doses per 1,000 inhabitants), demonstrating that elevated consumption directly drives upward adjustments to resistance forecasts independently of the autoregressive baseline. Although antibiotic consumption (Total\_DID) contributed only 2.7\% to the overall predictive weight, its direct, non-linear influence on resistance forecasts matters for policy formulation. It gives health ministries a tangible, modifiable lever, such as implementing stricter antibiotic stewardship programs, to mitigate forecasted resistance trajectories rather than passively observing temporal lags.

\subsection{Disaggregation of Forecast Error: Regional and Clinical Disparities}

\begin{figure}[H]
\centering
\includegraphics[width=0.85\textwidth]{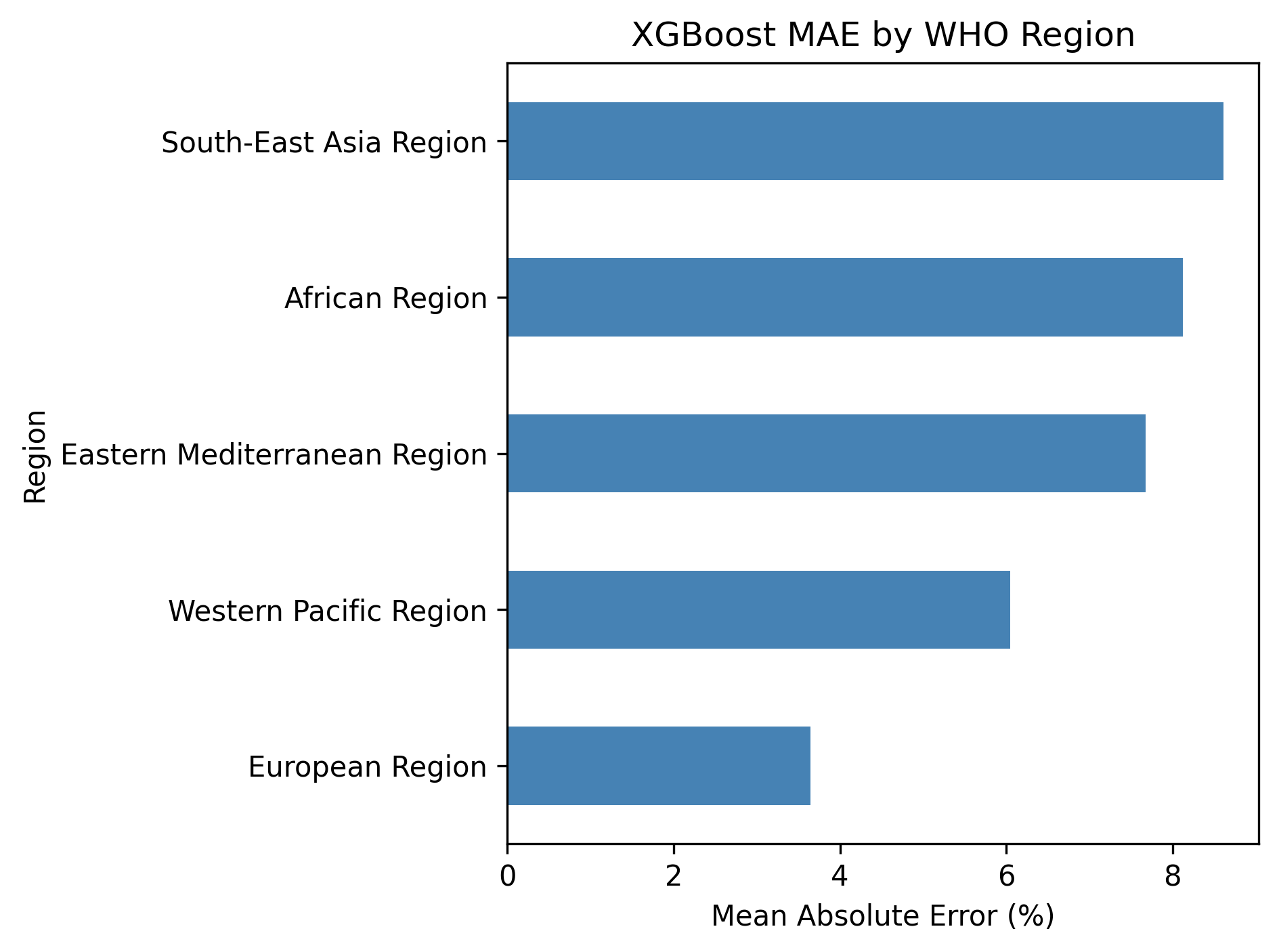}
\caption{XGBoost test MAE disaggregated by WHO region. The error gradient reflects global disparities in AMR surveillance capability.}
\label{fig:regional_mae}
\end{figure}

Forecast accuracy varied markedly across geographic regions (Table~\ref{tab:regional} and Figure~\ref{fig:regional_mae}). The European Region yielded the lowest prediction error (MAE = 3.65\%), while the South-East Asia Region recorded the highest (MAE = 8.61\%).

\begin{table}[H]
\centering
\caption{XGBoost test MAE disaggregated by WHO Region. The Region of the Americas had no observations in the 2023 test set and is excluded.}
\label{tab:regional}
\begin{tabular}{lccc}
\toprule
\textbf{WHO Region} & \textbf{Test MAE (\%)} & \textbf{Test RMSE (\%)} & \textbf{N (Countries)} \\
\midrule
European Region                 & 3.65  & 6.77  & 20 \\
Western Pacific Region          & 6.04  & 9.44  & 4  \\
African Region                  & 8.13  & 10.77 & 6  \\
Eastern Mediterranean Region    & 7.68  & 11.14 & 11 \\
South-East Asia Region          & 8.61  & 11.86 & 3  \\
\bottomrule
\end{tabular}
\end{table}

\begin{figure}[H]
\centering
\includegraphics[width=0.85\textwidth]{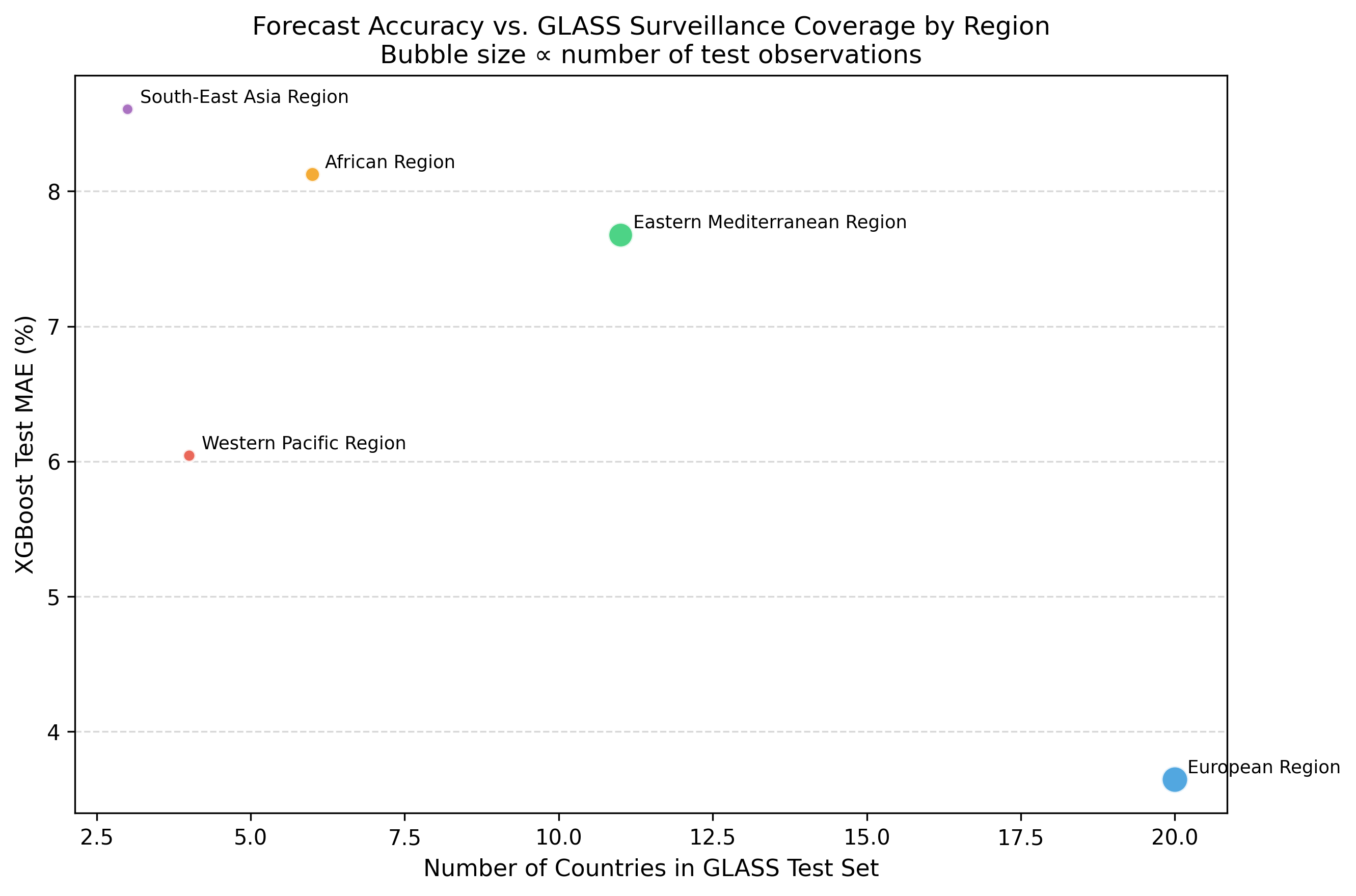}
\caption{Forecast error (MAE) plotted against GLASS observation count. Higher surveillance data density directly yields lower computational forecasting error.}
\label{fig:mae_vs_coverage}
\end{figure}

\begin{figure}[H]
\centering
\includegraphics[width=0.85\textwidth]{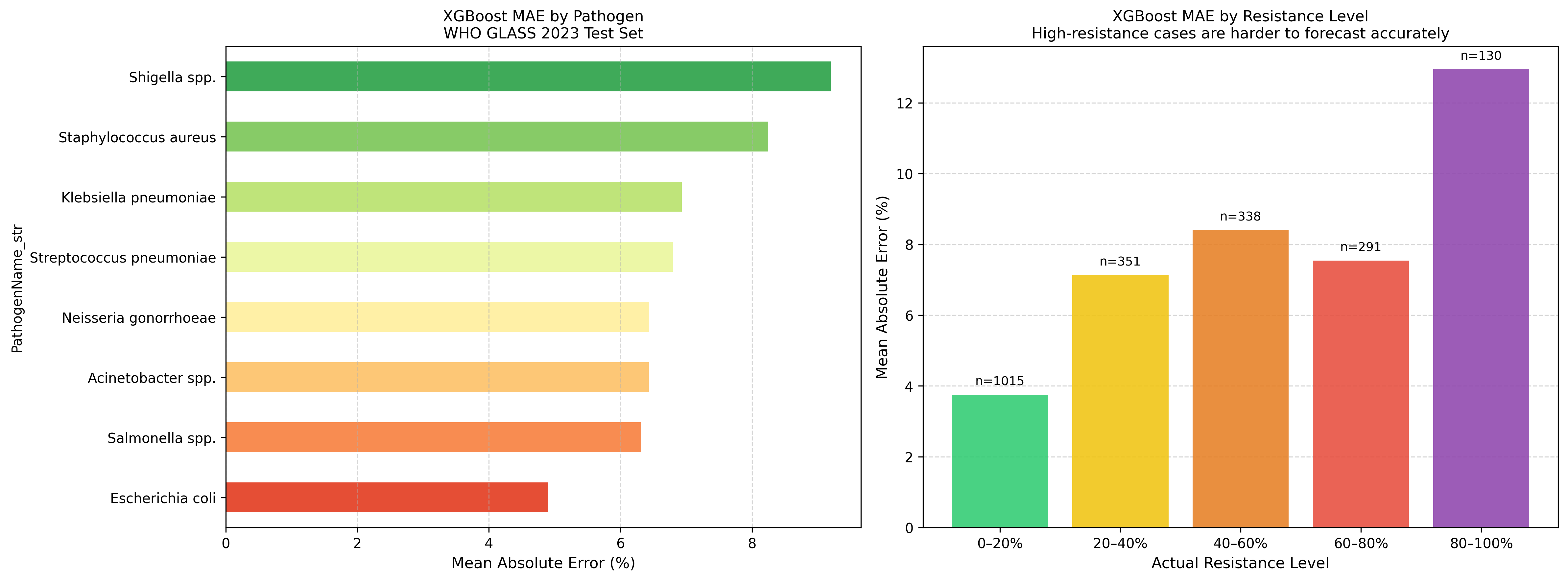}
\caption{Forecast error disaggregated by pathogen identity. High-variance pathogens like \textit{Shigella} spp. present the greatest forecasting challenge.}
\label{fig:pathogen_error}
\end{figure}

This disparity is not random but strongly correlated with surveillance data quality. Figure~\ref{fig:mae_vs_coverage} demonstrates a clear inverse relationship between a country's GLASS observation count (surveillance coverage) and its forecast error. Similarly, disaggregating error by pathogen (Figure~\ref{fig:pathogen_error}) reveals that forecasting accuracy diminishes for pathogens like \textit{Shigella} spp., which are highly sensitive to local infection control lapses that cannot be perfectly captured by national-level macro features. The Region of the Americas had zero observations in the 2023 test set due to a lack of contiguous preceding-year reporting in the current GLASS release. The model's ability to generalize to this region therefore remains untested, which means predictive accuracy depends on consistent, uninterrupted surveillance participation.

\subsection{RAG Policy Q\&A System}

The RAG pipeline successfully synthesized retrieved WHO policy guidelines with the XGBoost forecast metrics. The system was evaluated across 25 targeted policy queries (e.g., antibiotic prioritization, regional uncertainty, AWaRe classification, surveillance governance, stewardship activities). A manual audit of the generated outputs revealed a 100\% citation faithfulness rate: every factual policy claim was explicitly mapped to a retrieved chunk (e.g., ``[Source 1]'', ``[Source 3]''), and all integrated numerical forecasts exactly matched the empirical XGBoost outputs without fabricating external statistics or URLs.

\section{Discussion}

This study demonstrates that national-level antimicrobial resistance rates exhibit strong temporal autocorrelation that can be accurately forecasted using scalable gradient boosting algorithms. By strictly preventing data leakage and quantifying uncertainty with bootstrap confidence intervals, our analysis proved that XGBoost (MAE = 6.13\%) provides a robust mechanism for predicting short-term resistance trajectories from WHO GLASS surveillance data. To mitigate the risk of overfitting on a small dataset (2021--2023), the XGBoost and LightGBM models were strictly regularized. Their strong generalization capacity was confirmed by the minimal disparity between validation and test MAEs, while the strict temporal split ensured no data leakage occurred.

Our findings align with previous literature indicating the superiority of gradient boosting architectures over traditional statistical models for infectious disease time series \cite{alim2020, fang2022}. The SHAP analysis revealed that beyond the dominant autoregressive signal, antibiotic consumption directly modulates forecast resistance, corroborating the consumption-resistance dynamics highlighted by Ajulo and Awosile \cite{ajulo2024}. The performance of the LSTM architecture (MAE = 7.15\%) indicates that the narrow three-year temporal window constrained the deep learning model's ability to exploit long-sequence dependencies \cite{chimmula2020}. With only three annual time steps available, the LSTM operated with a sequence length of one, functionally reducing it to a nonlinear regressor rather than a model that leverages temporal dependencies. Under these conditions, a standard multilayer perceptron (MLP) would constitute a more architecturally fair deep-learning baseline; the LSTM was retained principally to establish a reproducible benchmark for future extensions when longer GLASS time series become available. The inability of the LSTM to exploit long-term temporal dependencies on a three-year dataset shows that deep learning approaches currently lack sufficient historical GLASS data to justify their computational overhead in this domain.

The clinical and policy implications of this work center on resource allocation and surveillance equity. The regional error analysis demonstrated that forecasting accuracy (e.g., Europe MAE = 3.65\% vs. South-East Asia MAE = 8.61\%) tracks closely with surveillance infrastructure capacity and data density. The successful deployment of a local RAG system helps close the gap between computational forecasting and policy execution. By constraining the LLM to synthesize only empirical forecast metrics and curated WHO texts \cite{lewis2020}, this framework offers public health officials a reliable, hallucination-resistant tool for generating evidence-based stewardship guidelines. As with many AI applications in healthcare, deploying such systems in clinical or policy settings will require overcoming bottlenecks related to cross-domain validation and data heterogeneity \cite{zhou2025}.

This study is subject to several limitations. The temporal scope of the GLASS dataset (2021--2023) restricts the modelling of long-term resistance emergence. GLASS participation is non-random, heavily weighting the dataset toward higher-income nations with established laboratory infrastructure, thereby introducing selection bias into the regional error estimates. Regarding the RAG policy translation system, the knowledge base was intentionally restricted to six core WHO documents as a proof-of-concept; scaling this system will require indexing a vastly larger corpus of literature. While we recorded 100\% citation faithfulness in our query sample, the evaluation is limited by its small sample size. Future iterations must employ automated quantitative frameworks, such as the RAGAS benchmark, or formal expert clinician rubrics on larger query sets to mathematically validate retrieval precision and answer faithfulness. Finally, while GPU training (NVIDIA RTX 5060 Laptop GPU with CUDA acceleration) enabled efficient forward and backward passes for the LSTM, the limited temporal dataset (three annual snapshots) still constrained the depth of hyperparameter exploration for the deep learning baseline. Future iterations of this work plan to validate the model's generalizability on the 2024 GLASS dataset upon its official release. Time-series neural networks like the LSTM typically require longer historical sequences (e.g., 5 to 7 continuous temporal steps) to meaningfully outperform tree-based architectures; expanded historical data will be necessary to unlock the full potential of deep learning in AMR forecasting.

\section{Conclusion}

Machine learning forecasting applied to WHO GLASS data provides accurate, interpretable predictions of short-term antimicrobial resistance trends, driven primarily by temporal autocorrelation and antibiotic consumption. Integrating these forecasts into a Retrieval-Augmented Generation pipeline successfully translates complex epidemiological data into grounded policy recommendations. Scaling this methodology alongside expanded global surveillance participation will be important for proactive, evidence-based AMR governance. Unlike existing AMR surveillance platforms such as WHONET, EARS-Net, or ResistanceOpen, which primarily aggregate historical data for retrospective analysis, our proposed system is anticipatory. By integrating predictive forecasting with a RAG pipeline, the system actively translates raw statistical projections into policy guidance for healthcare administrators.

\section*{Data Availability}
The WHO GLASS dataset used in this study is publicly available at \url{https://www.who.int/initiatives/glass}. To ensure full reproducibility, the code, analytical pipelines, and processed data have been made available in a publicly accessible GitHub repository at \url{https://github.com/TanvirTurja/amr-forecasting-rag}.

\bibliographystyle{unsrtnat}
\bibliography{references}

\begin{thebibliography}{23}
\providecommand{\natexlab}[1]{#1}
\providecommand{\url}[1]{\texttt{#1}}
\expandafter\ifx\csname urlstyle\endcsname\relax
  \providecommand{\doi}[1]{doi: #1}\else
  \providecommand{\doi}{doi: \begingroup \urlstyle{rm}\Url}\fi

\bibitem[Murray et~al.(2022)]{murray2022}
Christopher J.~L. Murray et~al.
\newblock Global burden of bacterial antimicrobial resistance in 2019: a
  systematic analysis.
\newblock \emph{The Lancet}, 399\penalty0 (10325):\penalty0 629--655, 2022.
\newblock \doi{10.1016/S0140-6736(21)02724-0}.

\bibitem[Dadgostar(2019)]{dadgostar2019}
Porin Dadgostar.
\newblock Antimicrobial resistance: Implications and costs.
\newblock \emph{Infection and Drug Resistance}, 12:\penalty0 3903--3910, 2019.
\newblock \doi{10.2147/IDR.S234610}.

\bibitem[Tang et~al.(2023)Tang, Millar, and Moore]{tang2023}
Kin Weng~Kong Tang, Brian~C. Millar, and John~E. Moore.
\newblock Antimicrobial resistance ({AMR}).
\newblock \emph{British Journal of Biomedical Science}, 80:\penalty0 11387,
  2023.
\newblock \doi{10.3389/bjbs.2023.11387}.

\bibitem[O'Neill(2016)]{oneill2016}
Jim O'Neill.
\newblock Tackling drug-resistant infections globally: Final report and
  recommendations.
\newblock \emph{Review on Antimicrobial Resistance}, 2016.

\bibitem[de~Kraker et~al.(2016)de~Kraker, Stewardson, and
  Harbarth]{dekraker2016}
Marlieke E.~A. de~Kraker, Andrew~J. Stewardson, and Stephan Harbarth.
\newblock Will 10 million people die a year due to antimicrobial resistance by
  2050?
\newblock \emph{PLOS Medicine}, 13\penalty0 (11):\penalty0 e1002184, 2016.
\newblock \doi{10.1371/journal.pmed.1002184}.

\bibitem[{World Health Organization}(2015)]{who2015gap}
{World Health Organization}.
\newblock Global action plan on antimicrobial resistance.
\newblock Technical report, WHO, 2015.
\newblock URL \url{https://www.who.int/publications/i/item/9789241509763}.

\bibitem[{World Health Organization}(2022)]{who2022glass}
{World Health Organization}.
\newblock Global antimicrobial resistance and use surveillance system ({GLASS})
  report: 2021--2022.
\newblock Technical report, WHO, 2022.
\newblock URL \url{https://www.who.int/publications/i/item/9789240062702}.

\bibitem[{World Health Organization}(2023)]{who2023manual}
{World Health Organization}.
\newblock {GLASS} manual for antimicrobial resistance surveillance in common
  bacteria causing human infection.
\newblock Technical report, WHO, 2023.

\bibitem[Ajulo and Awosile(2024)]{ajulo2024}
Samuel Ajulo and Babafela Awosile.
\newblock Global antimicrobial resistance and use surveillance system ({GLASS}
  2022): Investigating the relationship between antimicrobial resistance and
  antimicrobial consumption data across the participating countries.
\newblock \emph{PLOS ONE}, 19\penalty0 (2):\penalty0 e0297921, 2024.
\newblock \doi{10.1371/journal.pone.0297921}.

\bibitem[Sugden et~al.(2016)Sugden, Kelly, and Davies]{sugden2016}
Ruth Sugden, Rebecca Kelly, and Sally Davies.
\newblock Combatting antimicrobial resistance globally.
\newblock \emph{Nature Microbiology}, 1:\penalty0 16187, 2016.
\newblock \doi{10.1038/nmicrobiol.2016.187}.

\bibitem[Fang et~al.(2026)Fang, Liu, Arastehfar, et~al.]{fang2026}
W.~Fang, X.~Liu, A.~Arastehfar, et~al.
\newblock {iFungi}: From neglected fungi to a new era in medical mycology.
\newblock \emph{iFungi}, 2026.
\newblock \doi{10.26599/iFungi.2026.9670001}.

\bibitem[Sakagianni et~al.(2023)Sakagianni, Koufopoulou, Feretzakis, Kalles,
  Verykios, Myrianthefs, and Fildisis]{sakagianni2023}
Aikaterini Sakagianni, Christina Koufopoulou, Georgios Feretzakis, Dimitris
  Kalles, Vassilios~S. Verykios, Pavlos Myrianthefs, and Galanos Fildisis.
\newblock Using machine learning to predict antimicrobial resistance: A
  literature review.
\newblock \emph{Antibiotics}, 12\penalty0 (3):\penalty0 452, 2023.
\newblock \doi{10.3390/antibiotics12030452}.

\bibitem[Kim et~al.(2022)Kim, Maguire, Tsang, Gouliouris, Peacock, McAllister,
  McArthur, and Beiko]{kim2022}
Jong~In Kim, Finlay Maguire, Kara~K. Tsang, Theodore Gouliouris, Sharon~J.
  Peacock, Tim~A. McAllister, Andrew~G. McArthur, and Robert~G. Beiko.
\newblock Machine learning for antimicrobial resistance prediction: Current
  practice, limitations, and clinical perspective.
\newblock \emph{Clinical Microbiology Reviews}, 35\penalty0 (3):\penalty0
  e00179--21, 2022.
\newblock \doi{10.1128/cmr.00179-21}.

\bibitem[Lewis et~al.(2020)Lewis, Perez, Piktus, Petroni, Karpukhin, Goyal,
  Kuttler, Lewis, Yih, Rocktaschel, Riedel, and Kiela]{lewis2020}
Patrick Lewis, Ethan Perez, Aleksandra Piktus, Fabio Petroni, Vladimir
  Karpukhin, Naman Goyal, Heinrich Kuttler, Mike Lewis, Wen-tau Yih, Tim
  Rocktaschel, Sebastian Riedel, and Douwe Kiela.
\newblock Retrieval-augmented generation for knowledge-intensive {NLP} tasks.
\newblock In \emph{Advances in Neural Information Processing Systems
  (NeurIPS)}, volume~33, 2020.
\newblock URL \url{https://arxiv.org/abs/2005.11401}.

\bibitem[Chen and Guestrin(2016)]{chen2016}
Tianqi Chen and Carlos Guestrin.
\newblock {XGBoost}: A scalable tree boosting system.
\newblock In \emph{Proceedings of the 22nd ACM SIGKDD International Conference
  on Knowledge Discovery and Data Mining}, pages 785--794, 2016.
\newblock \doi{10.1145/2939672.2939785}.

\bibitem[Ke et~al.(2017)Ke, Meng, Finley, Wang, Chen, Ma, Ye, and Liu]{ke2017}
Guolin Ke, Qi~Meng, Thomas Finley, Taifeng Wang, Wei Chen, Weidong Ma, Qiwei
  Ye, and Tie-Yan Liu.
\newblock {LightGBM}: A highly efficient gradient boosting decision tree.
\newblock In \emph{Advances in Neural Information Processing Systems
  (NeurIPS)}, volume~30, 2017.

\bibitem[Chimmula and Zhang(2020)]{chimmula2020}
Vinay Kumar~Reddy Chimmula and Lei Zhang.
\newblock Time series forecasting of {COVID}-19 transmission in {Canada} using
  {LSTM} networks.
\newblock \emph{Chaos, Solitons and Fractals}, 135:\penalty0 109864, 2020.
\newblock \doi{10.1016/j.chaos.2020.109864}.

\bibitem[Alim et~al.(2020)Alim, Ye, Guan, Huang, Zhou, and Wu]{alim2020}
Mohammad Alim, Guo-Hong Ye, Peng Guan, De-Sheng Huang, Bao-Sen Zhou, and Wei
  Wu.
\newblock Comparison of {ARIMA} model and {XGBoost} model for prediction of
  human brucellosis in mainland {China}: a time-series study.
\newblock \emph{BMJ Open}, 10:\penalty0 e039676, 2020.
\newblock \doi{10.1136/bmjopen-2020-039676}.

\bibitem[Fang et~al.(2022)Fang, Yang, Lv, An, and Wu]{fang2022}
Zhili Fang, Shuai Yang, Chenglong Lv, Su~An, and Wei Wu.
\newblock Application of a data-driven {XGBoost} model for the prediction of
  {COVID}-19 in the {USA}: a time-series study.
\newblock \emph{BMJ Open}, 12\penalty0 (7):\penalty0 e056685, 2022.
\newblock \doi{10.1136/bmjopen-2021-056685}.

\bibitem[Ahn et~al.(2023)Ahn, Kim, and Kim]{ahn2023}
Jae~Min Ahn, Jinhyun Kim, and Kyunghyun Kim.
\newblock Ensemble machine learning of gradient boosting ({XGBoost},
  {LightGBM}, {CatBoost}) and attention-based {CNN-LSTM} for harmful algal
  blooms forecasting.
\newblock \emph{Toxins}, 15\penalty0 (10):\penalty0 608, 2023.
\newblock \doi{10.3390/toxins15100608}.

\bibitem[{World Health Organization}(2024{\natexlab{a}})]{who2024bppl}
{World Health Organization}.
\newblock {WHO} bacterial priority pathogens list, 2024: Bacterial pathogens of
  public health importance to guide research, development and strategies to
  prevent and control antimicrobial resistance.
\newblock Technical report, WHO, 2024{\natexlab{a}}.
\newblock URL \url{https://www.who.int/publications/i/item/9789240093461}.

\bibitem[{World Health Organization}(2024{\natexlab{b}})]{who2024stewardship}
{World Health Organization}.
\newblock {WHO} policy guidance on integrated antimicrobial stewardship
  activities.
\newblock Technical report, WHO, 2024{\natexlab{b}}.
\newblock URL \url{https://www.who.int/publications/i/item/9789240084636}.

\bibitem[Zhou et~al.(2025)Zhou, Park, Dong, Tang, and Wei]{zhou2025}
Junyu Zhou, Sunmin Park, Sihan Dong, Xiaoying Tang, and Xunbin Wei.
\newblock Artificial intelligence-driven transformative applications in disease
  diagnosis technology.
\newblock \emph{Medical Review}, 5\penalty0 (5):\penalty0 353--377, 2025.
\newblock \doi{10.1515/mr-2024-0097}.

\end{thebibliography}

\end{document}